%% file: main_camera_ready.tex
\documentclass[10pt, twocolumn, letterpaper]{article}

\usepackage{iccv}
\usepackage{times}
\usepackage{epsfig}
\usepackage{graphicx}
\usepackage{amsmath}
\usepackage{amssymb}

\usepackage{multirow}
\usepackage{xcolor}
\usepackage{wrapfig}
\usepackage{booktabs}
\usepackage{times}
\usepackage{makecell}
\usepackage{authblk}

\usepackage[breaklinks=true,bookmarks=false]{hyperref}

\iccvfinalcopy

\title{Technical Report for ICCV 2023 Visual Continual Learning Challenge:\\ 
Continuous Test-time Adaptation for Semantic Segmentation}

\makeatletter
\renewcommand\AB@affilsepx{, \protect\Affilfont}
\makeatother

\begin{document}

\author[1,2]{Damian~Sójka}
\author[3,4]{Yuyang~Liu}
\author[5,6]{Dipam~Goswami}
\author[2,7]{Sebastian~Cygert}
\author[2,5,6]{Bartłomiej~Twardowski}
\author[5,6]{Joost~van~de~Weijer}

\affil[1]{Poznań~University~of~Technology}
\affil[2]{IDEAS~NCBR}
\affil[3]{Shenyang~Institute~of~Automation}
\affil[4]{Chinese~Academy~of~Sciences}
\affil[5]{Computer~Vision~Center}
\affil[6]{Autonomous~University~of~Barcelona}
\affil[7]{Gdańsk~University~of~Technology}

\maketitle
\ificcvfinal\thispagestyle{empty}\fi

\section{Introduction}

The goal of the challenge is to develop a test-time adaptation (TTA) method, which could adapt the model to gradually changing domains in video sequences for semantic segmentation task. It is based on a synthetic driving video dataset - SHIFT~\cite{shift}. The source model is trained on images taken during daytime in clear weather. Domain changes at test-time are mainly caused by varying weather conditions and times of day. The TTA methods are evaluated in each image sequence (video) separately, meaning the model is reset to the source model state before the next sequence. Images come one by one and a prediction has to be made at the arrival of each frame. Each sequence is composed of 401 images and starts with the source domain, then gradually drifts to a different one (changing weather or time of day) until the middle of the sequence. In the second half of the sequence, the domain gradually shifts back to the source one. Ground truth data is available only for the validation split of the SHIFT dataset, in which there are only six sequences that start and end with the source domain. We conduct an analysis specifically on those sequences. Ground truth data for test split, on which the developed TTA methods are evaluated for leader board ranking, are not publicly available. 

The proposed solution secured a 3rd place in a challenge and received an innovation award. Contrary to the solutions that scored better, we did not use any external pretrained models or specialized data augmentations, to keep the solutions as general as possible. We have focused on analyzing the distributional shift and developing a method that could adapt to changing data dynamics and generalize across different scenarios.

\section{Problem Analysis}
We conduct an extensive analysis using the semantic segmentation source model DeepLabv3+~\cite{deeplabv3plus} with ResNet50~\cite{resnet} backbone. We utilize model weights provided by the challenge organizers. 

\begin{figure}[!htb]
    \centering
    \includegraphics[width=\columnwidth]{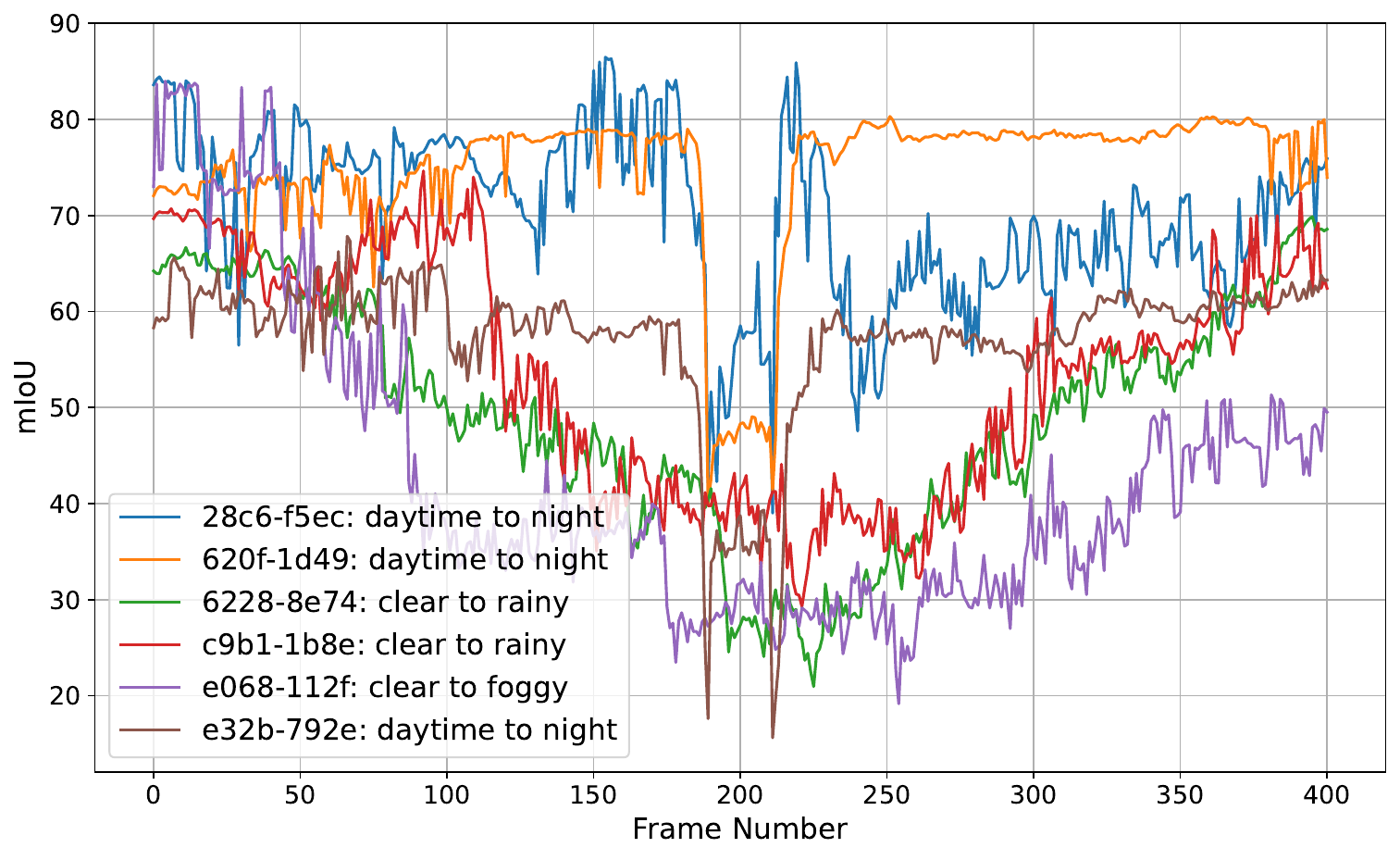}
    \caption{The plot of $mIoU$ between ground truth and source model predictions for each frame in six different sequences from the validation split.}
    \label{fig:source_miou}
    % \vspace{-4mm}
\end{figure}

Firstly, we check how the domain shift influences unchanged source model performance over the span of the sequence. Figure~\ref{fig:source_miou} shows the mean intersection over union (mIoU) for each image in six sequences from the validation split. As expected, the performance degrades until the middle of the sequence, where the domain shift is the most drastic, and starts increasing toward the end of the sequence, as the domain comes back to the source one. However, the change in mIoU is gradual for sequences in which the domain change is in the form of weather conditions (clear to rainy or foggy), and more abrupt for videos with domain change to night. It might be caused by drastic changes in lighting conditions during the night. This suggests that the developed method should be flexible and be able to adapt the model to both gradual and abrupt changes.

To further analyze the nature of domain shift, we decided to inspect the shift in data distribution $\phi=(\mu, \sigma)$, where $\mu$ is a mean of data and $\sigma$ represents a standard deviation. We examine the distance between the distributions of source data $\phi^S$ and test images $\phi_t^T$ for each frame at time $t$. We utilize symmetric Kullback-Leibler divergence as a distance metric $D(\phi^S, \phi_t^T)$:

\begin{equation}
\label{eq:sym_kl_div_for_plot}
    D(\phi^S, \phi_t^T) = \frac{1}{C}\sum_{i=1}^{C}KL(\phi_i^S || \phi_{t, i}^T) + KL(\phi_{t, i}^T || \phi_i^S)
\end{equation}
where $C$ is the number of channels.

\begin{figure}[!htb]
    \centering
    \includegraphics[width=\columnwidth]{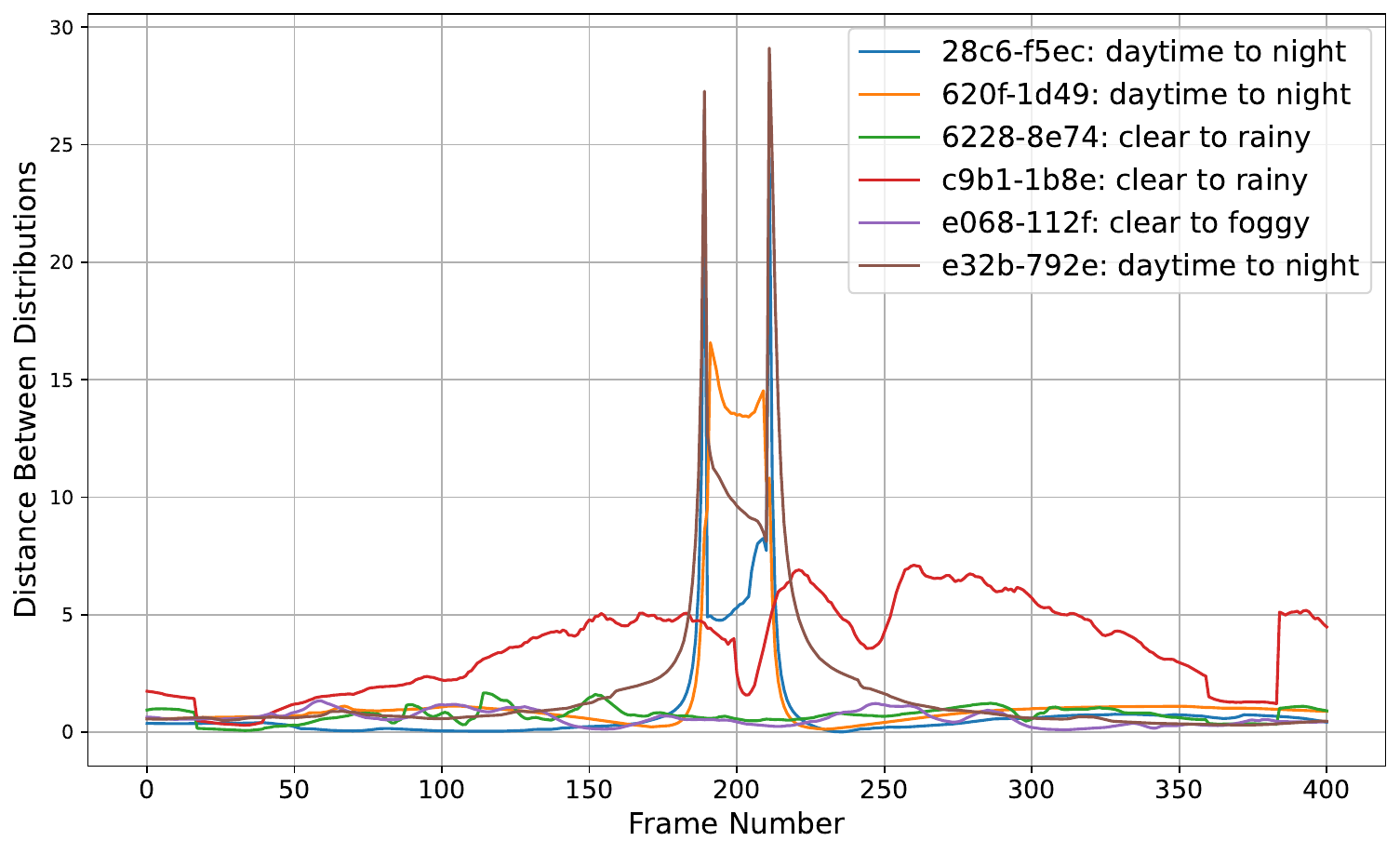}
    \caption{The plot of symmetric Kullback-Leibler divergence as a distance metric between source training data distribution and the distribution of each frame in six different sequences from the validation split.}
    \label{fig:kl_div_source_test}
    % \vspace{-4mm}
\end{figure}

The distance plot is presented in Figure~\ref{fig:kl_div_source_test}. It can be seen that the significance of the distribution shift varies between sequences. Changes in time of day to night time cause the test image distribution to greatly drift from the source data distribution. On the other hand, differences in weather conditions do not influence the distribution changes significantly. It shows that the TTA method should be able to handle different distributions of data and adjust the normalization process of the model accordingly, especially while using batch normalization (BN).

\begin{figure}[!htb]
    \centering
    \includegraphics[width=\columnwidth]{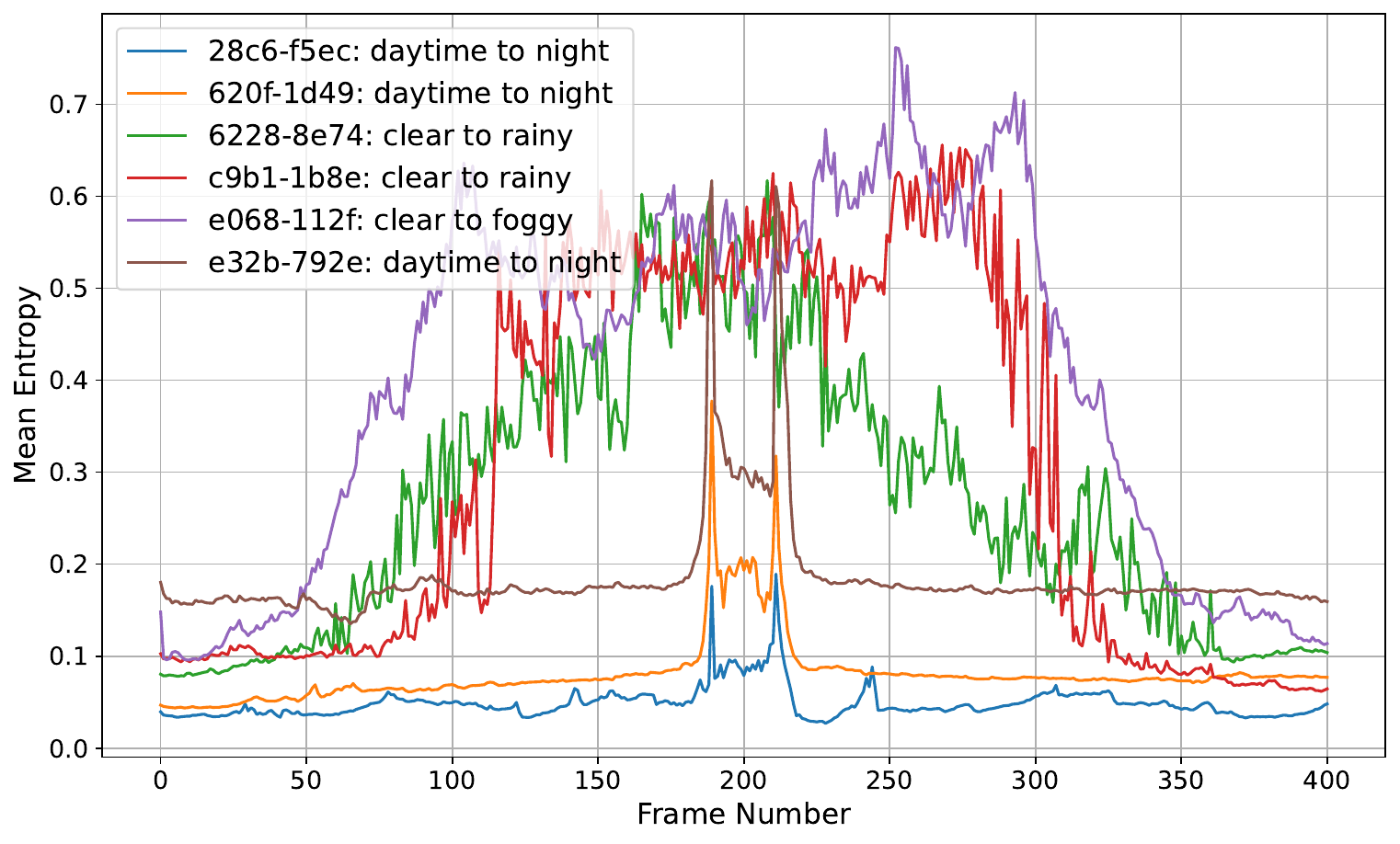}
    \caption{The plot of mean entropy of source model predictions for each frame in six different sequences from the validation split.}
    \label{fig:source_mean_entropy}
    % \vspace{-4mm}
\end{figure}

Lastly, Figure~\ref{fig:source_mean_entropy} shows the mean entropy of source model predictions for each frame of the sequences. Predictions' entropy increases with increasing domain shift. Moreover, the trend is highly similar to the $mIoU$ plot in Figure~\ref{fig:source_miou}. It suggests that entropy might be a relatively useful metric for evaluating the model performance and the degree of domain shift during test-time.

\section{Baselines}
There are two baseline TTA methods implemented by the challenge organizers: TENT~\cite{TENT} and CoTTA~\cite{COTTA}. They use two different adaptation methods. TENT uses prediction entropy minimization to update only batch normalization weights. CoTTA is based on adapting the student model with pseudo-labels generated by the teacher model. The teacher is updated by the exponential moving average of student's weights. To prevent performance degradation, it additionally uses stochastic model restoration, where randomly selected weights are reset to the source model state. Table~\ref{tab:baselines} shows their performance using local evaluation on six sequences from the validation split. DeepLabv3+~\cite{deeplabv3plus} model with weights provided by the organizers is used. TENT outperformed the CoTTA method in the challenge setting. Therefore, we choose entropy minimization used in TENT as our base adaptation method and build upon it.

\input{tables/baselines}

\section{Our Method}
Our method is composed of the base adaptation method from TENT~\cite{TENT} - entropy minimization, and three additional modifications. Firstly, considering the experimental results from Figure~\ref{fig:kl_div_source_test}, we utilize the dynamic BN statistics update method, described in Section~\ref{sec:dynamic_bn}. Moreover, to make the entropy minimization process more reliable, we filter the uncertain pixels from the minimization process by the value of the entropy of their prediction. We depict this approach in Section~\ref{sec:pixel_filtering}. Lastly, the original TENT method for classification tasks adapted only BN weights of the whole model. For segmentation, we only adapt the BN weights of the backbone (ResNet50), leaving the segmentation head fixed. We show the advantage of this approach experimentally in Section~\ref{sec:ablation}. 

\subsection{Dynamic Batch Normalization Statistics Update}
\label{sec:dynamic_bn}

Due to domain shift, state-of-the-art test-time adaptation methods~\cite{COTTA, eata, TENT} for classification task usually discard statistics calculated during source training and estimate data distribution based on each batch of data separately. However, this way of calculating the statistics is flawed since the sample size from data is usually too small to correctly estimate the data distribution, especially for lower batch sizes.

Moreover, as presented in Figure~\ref{fig:kl_div_source_test}, the magnitude of the distribution shift might vary between sequences. For some domain shifts, keeping the BN statistics of source data could be more beneficial. Therefore, there is a need for a method that adjusts statistics used in BN accordingly.

We adapt a part of the method developed by us~\cite{artta} to semantic segmentation task and use BN statistics from source data to estimate BN statistics $\phi_t = (\mu_t, \sigma_t)$ at time step $t$ during test-time by linearly interpolating between saved statistics from source data $\phi^S$ and calculated values from current batch $\phi^T_t$:
\begin{equation}
\label{eq:bn_interpolation}
    \phi_t = (1 - \beta)\phi^S + \beta \phi^T_t 
\end{equation}
where $\beta$ is a parameter that weights the influence of saved and currently calculated statistics. 

We utilize the symmetric KL divergence as a measure of distance between distributions $D(\phi_{t-1}, \phi_t^T)$ to adjust the value of $\beta$ accordingly to the severity of the distribution shift:
\begin{equation}
\label{eq:sym_kl_div}
    D(\phi_{t-1}, \phi_t^T) = \frac{1}{C}\sum_{i=1}^{C}KL(\phi_{t-1, i} || \phi_{t, i}^T) + KL(\phi_{t, i}^T || \phi_{t-1, i})
\end{equation}
where $C$ is the number of channels. $\beta_t$ at time step $t$ is calculated as follows:
\begin{equation}
\label{eq:new_beta}
    \beta_t = 1 - e^{-\gamma D(\phi_{t-1}, \phi_t^T)}
\end{equation}
where $\gamma$ is a scale hyperparameter.

To provide more stability for the adaptation, we take into account previous $\beta_{1:t-1}$ values and use an exponential moving average for $\beta_t$ update:
\begin{equation}
\label{eq:beta_ema}
    \beta = (1 - \alpha)\beta_{t-1} + \alpha \beta_t
\end{equation}
where $\alpha$ is a hyperparameter.

\subsection{Entropy-based Pixel Filtering}
\label{sec:pixel_filtering}
Training feedback from entropy minimization might be noisy and unreliable, considering that the model predictions can be incorrect. Some of the state-of-the-art TTA methods~\cite{eata, SAR} used for the classification task filter images on which they adapt based on the entropy of the model's predictions to make adaptation more reliable.  

We decided to use a similar approach for the task of segmentation. However, discarding the whole images on which we update the model could be sub-optimal, considering the low number of images in sequences and segmentation task. Instead, we mask out single pixels in the process of calculating the loss based on the entropy of prediction for those pixels. Pixels with predictions having an entropy higher than the predefined, constant threshold are masked and do not participate in the backpropagation process. This way, we are able to adapt more robustly, disregarding uncertain predictions.

\section{Experiments}

\subsection{Evaluation Metrics}
The main metrics defined by the organizers are as follows:
\begin{equation}
\label{eq:overall}
    overall = mIoU-2 \times mIoU_{drop}
\end{equation}
where $mIoU_{drop}$ is calculated as:
\begin{equation}
\label{eq:drop}
    mIoU_{drop} = mIoU_{source}-mIoU_{target}
\end{equation}
The $mIoU$ is based on combined predictions from all frames, $mIoU_{source}$ is a $mIoU$ from the first 20 frames of each sequence, and $mIoU_{target}$ is a $mIoU$ from 180th to 220th frame of each sequence.
Apart from $overall$, we utilize a simple $mIoU$ metric in our experiments.

\subsection{Implementation Details}

We use the code repository provided by the challenge organizers for the development and evaluation of our method. Additionally, we implemented the $overall$ metric locally ourselves, as it was only available on the evaluation server. The presented results are from our local evaluation on six sequences from the validation split unless stated otherwise. 
 
We utilize DeepLabv3+~\cite{deeplabv3plus} with ResNet50~\cite{resnet} backbone as a source model, with weights provided by the organizers. During TTA, we use a learning rate equal to 0.00006/4. The $\gamma$ parameter value from Equation~\ref{eq:new_beta} is set to 0.1 and $\alpha$ from Equation~\ref{eq:beta_ema} to 0.005, unless stated otherwise. The entropy threshold for discarding the pixels with uncertain predictions from the adaptation process is equal to $0.3 \times \ln{14}$, where 14 is the number of classes and the $\ln{14}$ represents the maximum entropy value.
 
\subsection{Results}
Table~\ref{tab:each_component} presents the performance with different combinations of components of our method. It can be seen that adding each element to our base method (entropy minimization) increases both $mIoU$ and overall metrics. The most significant improvement is achieved in terms of overall metric by adding Dynamic BN Statistics Update to \textbf{B} configuration.

\input{tables/each_component}

We show our final results from the server evaluation on test split in Table~\ref{tab:test_results}.

\input{tables/test_results}

\subsection{Ablation Study}
\label{sec:ablation}

Table~\ref{tab:adapted_parts} displays the performance of the baseline TENT~\cite{TENT} method, while different parts of a model are updated during test-time. The results show that adapting only BN weights of the backbone achieves the best results in terms of both metrics. Moreover, keeping all the weights of an adapted part unfrozen, instead of only BN ones, significantly degrades the plain entropy minimization performance.

\input{tables/adapting_different_parts}

Additionally, we explored different thresholds for filtering the unreliable pixel predictions for adaptation. Results are displayed in Table~\ref{tab:entropy_thresholds}.

\input{tables/entropy_thresholds}

Lastly, in Table~\ref{tab:dynamic_bn_params}, we show the performance of our method with different parameters of dynamic BN statistics update, namely $\gamma$ from Equation~\ref{eq:new_beta} and $\alpha$ from Equation~\ref{eq:beta_ema}.

\input{tables/dynamic_bn_params}

\subsection{Things That Didn't Work}
Apart from the techniques used in our final method, we experimented with more approaches that didn't work. 

We tried to adapt the model during test-time while keeping the features of the original and augmented images consistent by adding additional term to the loss function. 

Moreover, we explored preserving a buffer of a low number of previous predictions and averaging them to obtain more reliable pseudo-labels. Additionally, to account for different positions of objects in images in between the frames, we considered using optical flow to unify the predictions into a single time step. 

\section{Conclusions}
In this work, we present a brief analysis of the problem of continuous test-time adaptation and demonstrate our methods for semantic segmentation task. We were able to build upon and outperform the baselines.

\small{
\vspace{1ex}
\noindent\textbf{Acknowledgement.}
Bartłomiej Twardowski acknowledges the grant RYC2021-032765-I.
}

{\small
\bibliographystyle{ieee_fullname}
\bibliography{bibliography}
}

\end{document}

%% file: tables/baselines.tex
\begin{table}[h!]
\centering
\caption{mIoU and overall metrics of fixed source model, TENT~\cite{TENT} and CoTTA~\cite{COTTA} baselines. The results are from the local evaluation on the validation split. The learning rate is equal to 0.00006/8.}
\label{tab:baselines}
\scalebox{1.0}{
\begin{tabular}{c|cc}
\hline

Method & $mIoU$ & $overall$ \\ \hline

Source model & 54.4 & 12.9 \\
TENT~\cite{TENT} & \textbf{58.1} & \textbf{41.0} \\
CoTTA~\cite{COTTA} & 56.2 & 12.0 \\

\hline
\end{tabular}}
\end{table}

%% file: tables/each_component.tex
\begin{table}[htb!]
\centering
\caption{$mIoU$ and $overall$ metrics of different combinations of components of our method. The results are from the local evaluation on the validation split.}
\label{tab:each_component}
\scalebox{0.85}{
\begin{tabular}{l|cc}
\hline

Method & $mIoU$ & $overall$ \\ \hline

\textbf{A}: TENT~\cite{TENT} baseline (entropy minimization) & 56.9 & 39.0 \\

\textbf{B}: \textbf{A} + Adapting Only Backbone's BN Weights & 57.7 & 41.3 \\

\textbf{C}: \textbf{B} + Dynamic BN Statistics Update & 58.6 & 49.1 \\

\textbf{Ours}: \textbf{C} + Pixel Filtering  & \textbf{58.8} & \textbf{50.2} \\

\hline
\end{tabular}}
\end{table}

%% file: tables/test_results.tex
\begin{table}[h!]
\centering
\caption{Evaluation of the performance of our method from the evaluation server on the test split.}
\label{tab:test_results}
\scalebox{0.9}{
\begin{tabular}{cccc|c}
\hline

$mIoU$ & $mIoU_{drop}$ & $mIoU_{source}$ & $mIoU_{target}$ & $overall$ \\ \hline

71.4 & 23.3 & 76.5 & 53.2 & 24.7 \\

\hline
\end{tabular}}
\end{table}

%% file: tables/adapting_different_parts.tex
\begin{table}[h!]
\centering
\caption{$mIoU$ and $overall$ metrics of TENT~\cite{TENT} baseline with different configurations of adapted weights. The results are from the local evaluation on the validation split.}
\label{tab:adapted_parts}
\scalebox{0.85}{
\begin{tabular}{ccc|cc}
\hline

\multicolumn{3}{c|}{Adapted Weights} & \multirow{2}{*}{$mIoU$} & \multirow{2}{*}{$overall$} \\ \cline{0-2}
Backbone & Head & BN weights only & & \\ \hline

% FOR LR = 0.00006/8

% \checkmark &  &  & 7.7 & -108.3 \\  
%  & \checkmark &  & 26.8 & -44.2 \\

% \checkmark & & \checkmark & \textbf{58.6} & \textbf{42.1} \\  
%  & \checkmark & \checkmark & 54.4 & 14.7 \\

% \checkmark & \checkmark & \checkmark & 58.1 & 41.0 \\

% FOR LR = 0.00006/4

\checkmark &  &  & 4.6 & -101.4 \\
 & \checkmark &  & 18.7 & -61.8 \\

 \checkmark & \checkmark &  & 4.4 & -88.3 \\

\checkmark & & \checkmark & \textbf{57.7} & \textbf{41.3} \\  
 & \checkmark & \checkmark & 54.4 & 16.3 \\

\checkmark & \checkmark & \checkmark & 56.9 & 39.0 \\

\hline
\end{tabular}}
\end{table}

%% file: tables/entropy_thresholds.tex
\begin{table}[h!]
\centering
\caption{Performance of our method with different values of entropy threshold for filtering the unreliable pixel predictions. The results are from the local evaluation on the validation split.}
\label{tab:entropy_thresholds}
\scalebox{1.0}{
\begin{tabular}{c|cc}
\hline

Entropy threshold & $mIoU$ & $overall$ \\ \hline

$0.40 \times \ln{14}$ & 58.6 & 49.2 \\
$0.35 \times \ln{14}$ & 58.6 & 48.9 \\
$0.30 \times \ln{14}$ & \textbf{58.8} & \textbf{50.2} \\
$0.25 \times \ln{14}$ & \textbf{58.8} & 50.0 \\
$0.20 \times \ln{14}$ & \textbf{58.8} & 49.4 \\
$0.15 \times \ln{14}$ & 58.7 & 47.5 \\

\hline
\end{tabular}}
\end{table}

%% file: tables/dynamic_bn_params.tex
\begin{table}[h!]
\centering
\caption{Performance of our method with different values of dynamic BN statistics update parameters - $\gamma$ and $\alpha$. The results are from the local evaluation on the validation split.}
\label{tab:dynamic_bn_params}
\scalebox{1.0}{
\begin{tabular}{cc|cc}
\hline

$\gamma$ & $\alpha$ & $mIoU$ & $overall$ \\ \hline

\multirow{4}{*}{1} 
& 0.5 & 58.0 & 40.3 \\
& 0.05 & 58.3 & 42.0 \\
& 0.005 & 58.3 & 43.8 \\
& 0.0005 & 58.6 & 46.0 \\ 

\hline

\multirow{4}{*}{0.1} 
& 0.5 & 58.2 & 45.0 \\
& 0.05 & 58.7 & 48.2 \\
& 0.005 & \textbf{58.8} & \textbf{50.2} \\
& 0.0005 & \textbf{58.8} & 47.5 \\ 

\hline

\multirow{4}{*}{0.01} 
& 0.5 & 58.2 & 43.8 \\
& 0.05 & 58.4 & 45.6 \\
& 0.005 & 58.6 & 49.6 \\
& 0.0005 & \textbf{58.8} & 47.9 \\ 

\hline

\multirow{4}{*}{0.001} 
& 0.5 & 58.1 & 43.3 \\
& 0.05 & 58.3 & 45.2 \\
& 0.005 & \textbf{58.8} & 48.0 \\
& 0.0005 & \textbf{58.8} & 48.0 \\ 

\hline

\hline
\end{tabular}}
\end{table}